# Enhancing sampling-based planning with a library of paths

Michal Minařík *, Vojtěch Vonásek, Robert Pěnička

*Department of Cybernetics, Faculty of Electrical Engineering, Czech Technical University in Prague, Technická 2, Prague, 166 27, Czech Republic*

ABSTRACT

Path planning for 3D solid objects is a challenging problem, requiring a search in a six-dimensional configuration space, which is, nevertheless, essential in many robotic applications such as bin-picking and assembly. The commonly used sampling-based planners, such as Rapidly-exploring Random Trees, struggle with narrow passages where the sampling probability is low, increasing the time needed to find a solution. In scenarios like robotic bin-picking, various objects must be transported through the same environment. However, traditional planners start from scratch each time, losing valuable information gained during the planning process. We address this by using a library of past solutions, allowing the reuse of previous experiences even when planning for a new, previously unseen object. Paths for a set of objects are stored, and when planning for a new object, we find the most similar one in the library and use its paths as approximate solutions, adjusting for possible mutual transformations. The configuration space is then sampled along the approximate paths. Our method is tested in various narrow passage scenarios and compared with state-of-the-art methods from the OMPL library. Results show significant speed improvements (up to 85% decrease in the required time) of our method, often finding a solution in cases where the other planners fail. Our implementation of the proposed method is released as an open-source package.

## 1. Introduction

The task of path planning for a solid 3D object is to find a feasible (collision-free) path in an environment with obstacles, which arises in many applications in robotics and other fields, such as autonomous car navigation [1], CAD design [2,3], and assembly [4]. This leads to a search in six-dimensional configuration space, which can be solved using sampling-based approaches [5] like Rapidly-exploring Random Trees (RRT) [6] or Probabilistic Roadmaps (PRM) [7]. Sampling-based planners randomly sample the configuration space and maintain a graph structure (roadmap) of the collision-free samples. A solution (path) is found in the roadmap using a graph-based search. The narrow passage problem arises when the solution leads through a narrow collision-free region whose volume is significantly smaller than the volume of the whole configuration space. In such a case, the probability of placing samples into the narrow passage is low. Consequently, many samples must be drawn to find a solution through the narrow passage, which increases planning runtime. The narrow passage problem has been studied in many papers surveyed in [8], resulting in various approaches to cope with the problem. Nevertheless, the planning for a solid 3D object in the presence of a narrow passage is still an open problem as both the success rate (of finding a feasible path) and the computational time of the existing methods are greatly affected, as shown in this paper.

In guided sampling [9–12], a guide (a path in the workspace or even in the configuration space) is used to generate the random samples along it. A guiding path in the configuration space can be found by solving a similar, yet simpler, problem [13–16]. In scenarios requiring path planning for different objects in the same environment (e.g., in bin-picking [17,18]), computing the guiding paths prior to every planning would introduce a considerable overhead and slow the planning process.

This paper proposes a novel method named *Rapidly-exploring Random Trees with a Library of Paths* (RRT-LIB), shown in Fig. 1. We aim to reuse the knowledge gained by previous planning and generalize it to different objects moving through the same environment. In the *preparation phase*, we compile a library containing paths for multiple object classes. The paths are generated using a planner that is able to quickly approximate paths through an environment with narrow passages. The planner then leverages this knowledge in the *planning phase*. When a path for a new object is required, we retrieve the paths of the most similar object in the library. We choose the most similar object using a state-of-the-art 3D shape similarity evaluation method [19]. The relative transformation between the library and manipulated objects is computed using the Iterative Closest Point algorithm (ICP) [20]. The library paths are then transformed accordingly to be more relevant for the current planning task. Finally, these transformed paths serve as guiding paths, hinting at the possible paths through the environment.

* Corresponding author.
*E-mail addresses:* michal.minarik@fel.cvut.cz (M. Minařík), vonasek@labe.felk.cvut.cz (V. Vonásek), penicrob@fel.cvut.cz (R. Pěnička).

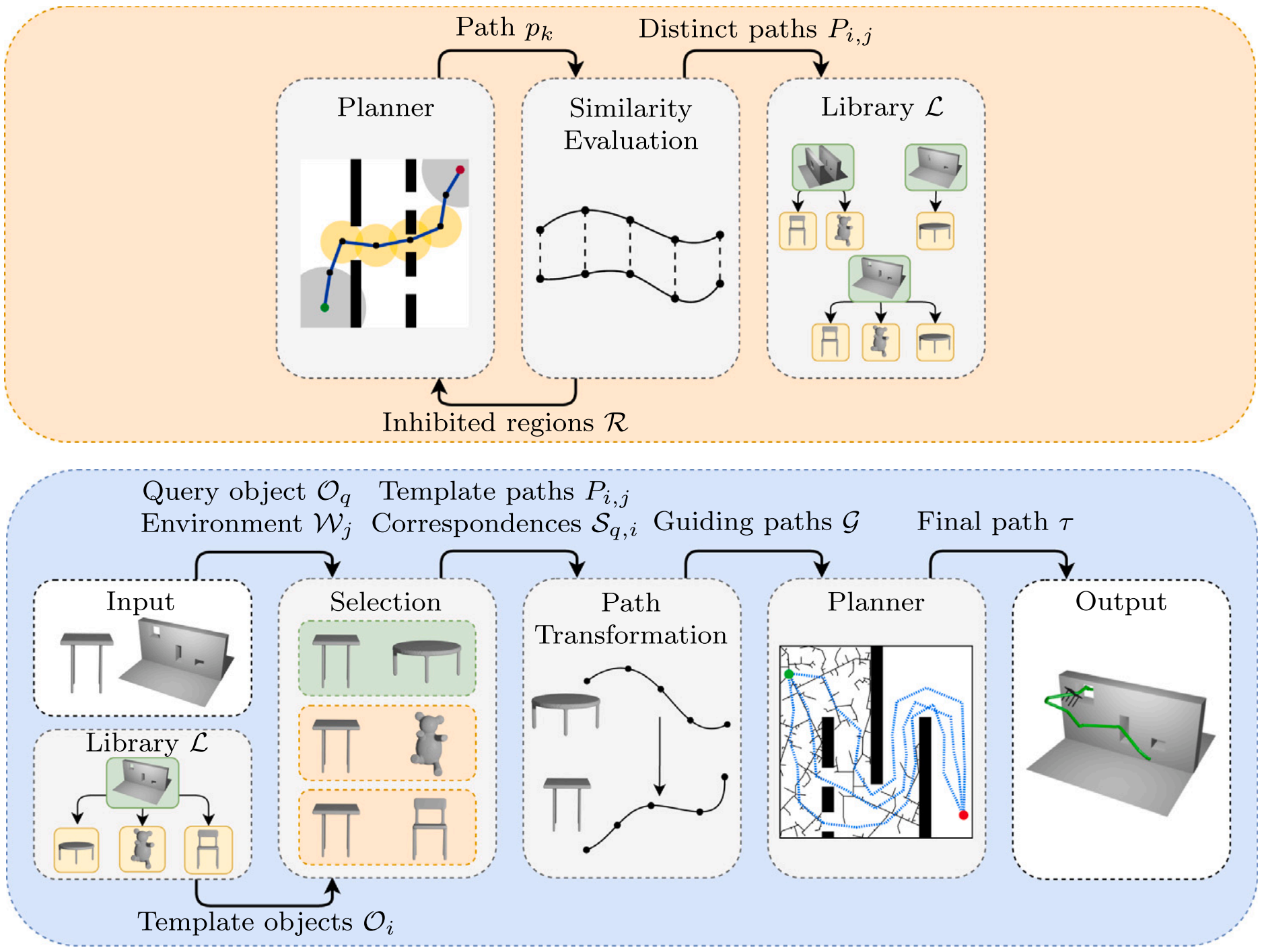

**Fig. 1.** RRT-LIB: *Rapidly-exploring Random Trees with a Library of Paths* — a novel algorithm for path planning in environments with narrow passages, consisting of the *preparation phase* (top) and the *planning phase* (bottom).

We demonstrate the effectiveness of our method on various environments and objects using a benchmarking tool inside the Open Motion Planning Library (OMPL) [21]. We show that RRT-LIB outperforms the existing methods in various environments containing multiple narrow passages, offering up to 85% decrease in the time needed and often being the only planner able to find a solution in the given amount of time.

## 2. Related work

The most well-known sampling-based planners are Rapidly-exploring Random Trees (RRT) [6] and Probabilistic Roadmaps (PRM) [7]. Several planners were derived from basic RRT and PRM, such as their asymptotically optimal variants RRT* and PRM* [22]. We refer to the surveys [8,23–25] about general variants of sampling-based planners. The survey [26,27] focuses on sampling under constraints and surveys [28,29] on optimal path planning.

Sampling-based planners can suffer from the well-known narrow passage problem [30,31], where a solution leads through a relatively small collision-free region of the configuration space. Due to its volume (relative to the volume of the whole configuration space), the probability of placing enough samples into the narrow passage is low. This increases the number of samples (and therefore the time) the planners need to find a solution.

Early attempts to cope with the narrow passages increase the probability of generating random samples around the obstacles. For example, the Gaussian PRM [32] generates two close samples. If exactly one of them is collision-free, it is added to the roadmap, effectively increasing the density of the roadmap near the obstacles. The strategy [32] was further extended in the Bridge-test PRM [33].

Among the basic extensions of RRT to cope with the narrow passage problem is the bidirectional RRT, where two trees are built simultaneously [34] or RRT with multiple trees [35]. However, the above-mentioned planners [32–35] do not utilize the knowledge of the workspace (e.g., shapes of the obstacles or information about the medial axis of the workspace).

Utilizing workspace knowledge can significantly improve planning performance. The XXL planner [36] utilizes a decomposed workspace to generate samples for various control points on a many-DOF robot. The MAPRM [37] approach retracts the samples towards the medial axis of the workspace. Similarly, the RRT-based planner [38] utilizes the medial axis of the workspace. RRV [39] uses collision-free and colliding samples to identify the type of local space (e.g., an entrance to a narrow passage). It uses PCA (Principal Component Analysis) to analyze the type of local space. In [40], a machine learning-based approach is used to estimate new relevant regions for sampling. The learning uses the samples and the information about collision detection and cost-to-go heuristics collected during the sampling.

A large family of methods utilizes the concept of guided-based planning, where the random samples are generated along a guide (path) instead of the whole configuration space. In case the environment contains a narrow passage and a guiding path going through the narrow passage is available, using the guide increases the probability of sampling in the narrow passage, thereby decreasing the time needed to find a solution. The simple guide can be represented as a path in the 2D workspace [11,41], a path in the 3D workspace [42], or a medial axis [38]. The work [43] utilizes multiple guides for motion planning of a mobile manipulator. One guiding path computed on the ground is used for generating samples for the mobile platform, and planning for the manipulator is guided using a 3D path generated in the workspace. The work [44] uses bidirectional search to sample along waypoints of previously found paths.

The performance of sampling along a workspace-based guide decreases with the increasing dimension of the configuration space [10]. Therefore, other techniques must be employed to obtain the guide for high-dimensional spaces. Researchers in [13,15,16,42,45,46] proposed computing the guide directly in the configuration space. This is achieved by solving a similar but simpler (relaxed) problem. For example, collision constraints can be relaxed by scaling-down the robot (or obstacles) [16] or by their thinning [47]. After the approximate solution is found, the configuration space is sampled again with the original robot.

In a mostly static workspace (e.g., mobile robots used for shelf stacking in a warehouse), experience can enhance planning performance. The Experience-Driven Random Trees (ERT) [48] method saves solution paths into a database as the robot completes various planning tasks. The path segments represent the robot's experience and are used to expand the search tree in subsequent planning tasks. In [49], the experience is a movement through a specific obstacle region (e.g., a robot's arm moving through a bookcase shelf). A standard sampling-based planner is used to find a path during the first interaction with a local obstacle region. This path is saved to a database, and when a similar obstacle region is encountered again, the database path is used to guide the planner. However, this approach requires an explicit model of the obstacles to create the local primitives. In both approaches, the experience depends on both the robot and the workspace. Therefore, the methods are suitable for situations where one robot repeatedly works in the same environment. In [50], a database of previously found paths for a robotic manipulator is used in parallel with planning-from-scratch. The path from the database is selected based on the distance of the start and goal configurations, and repaired if necessary. The repairing process identifies all valid segments of the path and attempts to connect them using bidirectional RRT. Simultaneously, planning from scratch aims to find new paths in the configuration space. A path found by either the database-based planner or by the planning-from-scratch is executed on the robot, and the database is updated if the new paths differ from the stored ones. The framework [50] is further extended in [51], but instead of storing whole paths, previous experience is stored in sparse roadmap spanners [52]. Sparse roadmap spanners can represent known connectivity in large configuration spaces and are less memory demanding than storing all previously found paths. The solution database is also used in [44]; each path is associated with a situation descriptor (e.g., information about start, goal, distance from obstacles, shape of obstacles around the path) and the descriptor is used to find a matching path from the database. The configuration space in [44] is searched using bidirectional RRT with dense sampling along the waypoints of the matching path. Unlike our method proposed in this paper, the framework [44] does not automatically decide if the newly found solution is stored in the database; it is decided by the user.

While guided-based planners [10–13,15,16,42,45] and database-based approaches [44,49–51] use stored paths to sample the configuration space (e.g., sampling along the paths), the approach [53] uses the path database to compute the heuristic determining which nodes of the search tree should be expanded. Moreover, the database is updated during the search (and not only after a valid path is found as in [44,50,51]).

Learning-based approaches are also presented in the available literature. In Motion Planning Networks (MPNet), two neural networks are used to plan the path incrementally [54]. The first network encodes the environment with obstacles into a latent space. The other network then takes this encoding, the current robot configuration, and the goal configuration and outputs the following configuration on the path. An optimal planner such as RRT* is used to teach the PNet when the neural network cannot provide a feasible trajectory. The drawback is that the same robot needs to stay the same between the training and the online execution. In [55], a Convolutional Neural Network is trained to predict sampling distribution for RRT*. The network is trained on more than $10^6$ examples of optimal trajectories computed by A*. The approach [55] is, however, limited only to 2D configuration spaces (as the input to the network is a 2D image of the workspace) and can work only for circular robots but not for general shapes as in our method. In [56], an autoencoding network is used to learn the latent space of the configuration space (the training data contains short feasible trajectories), and the planning is then realized using the latent space. The training requires many examples (authors use $10^4$ trajectories).

Such training by examples is similar to Learning from Demonstration [57], which leverages motions generated by a human expert when solving new problem instances (e.g., a robotic arm reproducing several tasks, such as grasping, placing, and releasing [58]).

The most relevant approaches for our method are [44,49–51,53]. While these frameworks assume only one type of robot, we consider multiple types of robots (objects). After a matching (guiding) path is found in the database for a given robot (object), it is not directly used to sample the configuration space like in [44,49], but it is first transformed according to the similarity between the object in the database and the current one. Similarly to the related works, our planner can update the library after discovering a new path. However, the library is updated only if the new path is distinct from already stored ones.

## 3. Rapidly-exploring random trees with a library of paths

The proposed method aims to solve path planning for various (query) 3D objects moving in a set of $m$ environments. The objects are allowed to move and rotate in the environment — therefore, their state is described by a configuration $q \in SE(3)$. Instead of searching the configuration space for each object from scratch, we build a library of template paths for a set of $n$ template objects in each environment. Then, to find a path for a query object $\mathcal{O}_q$, the most similar object in the library is found first and the template paths for it are retrieved. These paths are considered as approximate solutions for the query object, and the configuration space is sampled densely along them. Let $\mathcal{O}_i, i \in \{1,\dots,n\}$ be the template objects and $\mathcal{W}_j, j \in \{1,\dots,m\}$ the environments each described by a 3D triangular mesh. To compute the template paths, a scaled-down version $o_i$ of the template object $\mathcal{O}_i$ is used in order to promote finding multiple approximate guiding paths, as discussed in [14,45]. Each pair of an object $o_i$ and an environment $\mathcal{W}_j$ forms its own configuration space $\mathcal{C}_{i,j}$. For the sake of simplicity, we denote the configuration space as $\mathcal{C}$ and its collision-free region as $\mathcal{C}_{\text{free}} \subseteq \mathcal{C}$. The library $\mathcal{L} = (\mathcal{O}_i, \mathcal{W}_j, P_{i,j} \in \mathcal{C}_{\text{free}})$ contains $k_{i,j}$ template paths $P_{i,j} = \{p_1,\dots,p_{k_{i,j}}\}$ for the scaled-down version $o_i$ of the $i$th template object $\mathcal{O}_i$ in the $j$th environment. The path $p_k = (q_l), q_l \in \mathcal{C}_{\text{free}}$ is a sequence of collision-free waypoints. The ultimate goal of the method is to solve planning from the start towards the goal configuration in an environment $\mathcal{W}_j$. We assume that all template objects are collision-free when placed at both start and goal configurations in the environment $\mathcal{W}_j$. The proposed RRT-LIB method consists of two main phases: the preparation phase and the planning phase.

### 3.1. Preparation phase

The purpose of the preparation phase is to create a library of paths $\mathcal{L} = \{P_{i,j}\}$, where $P_{i,j}$ is a set of distinct paths for a template object $\mathcal{O}_i$ and an environment $\mathcal{W}_j$. The preparation phase is depicted in Fig. 2. This phase simulates a process of collecting plans from the past planning experience, when an entirely new type of manipulated object is introduced to the planning and the user decides it is worth creating a new template object instead of using the templates already available in the database. Alternatively, the library $\mathcal{L}$ can be updated during the planning phase every time a new path distinct from the already stored ones is found. Based on the similarity metric, the algorithm can even automatically decide to create a new template object. These possible extensions are discussed in Section 5. A naive approach to creating the library would be to employ a standard sampling-based planner (such as RRT) and repeatedly find various paths. However, sampling-based planners usually prefer to find easily achievable paths [59], and repeated planning from $q_{\text{start}}$ to $q_{\text{goal}}$ might result in similar paths with the same homotopy (i.e., paths that can be continuously deformed from one to the other without passing through the obstacle region) [5].

To promote finding non-similar paths, we employ the RRT-IR (RRT with Inhibited Regions) planner [13], that is able to find multiple distinct paths in the environment. RRT-IR repeatedly searches the configuration space, and in each planning trial attempts to discover a new path. The configuration space is sampled uniformly, but the planner avoids certain "prohibited" regions $\mathcal{R} \subset \mathcal{C}$. In the beginning, the whole configuration space can be searched (i.e., $\mathcal{R} = \emptyset$). After a path is found, its waypoints are added into $\mathcal{R}$, so that the subsequent search will avoid sampling along the previously found paths. The preparation phase of RRT-LIB is presented in Alg. 1.

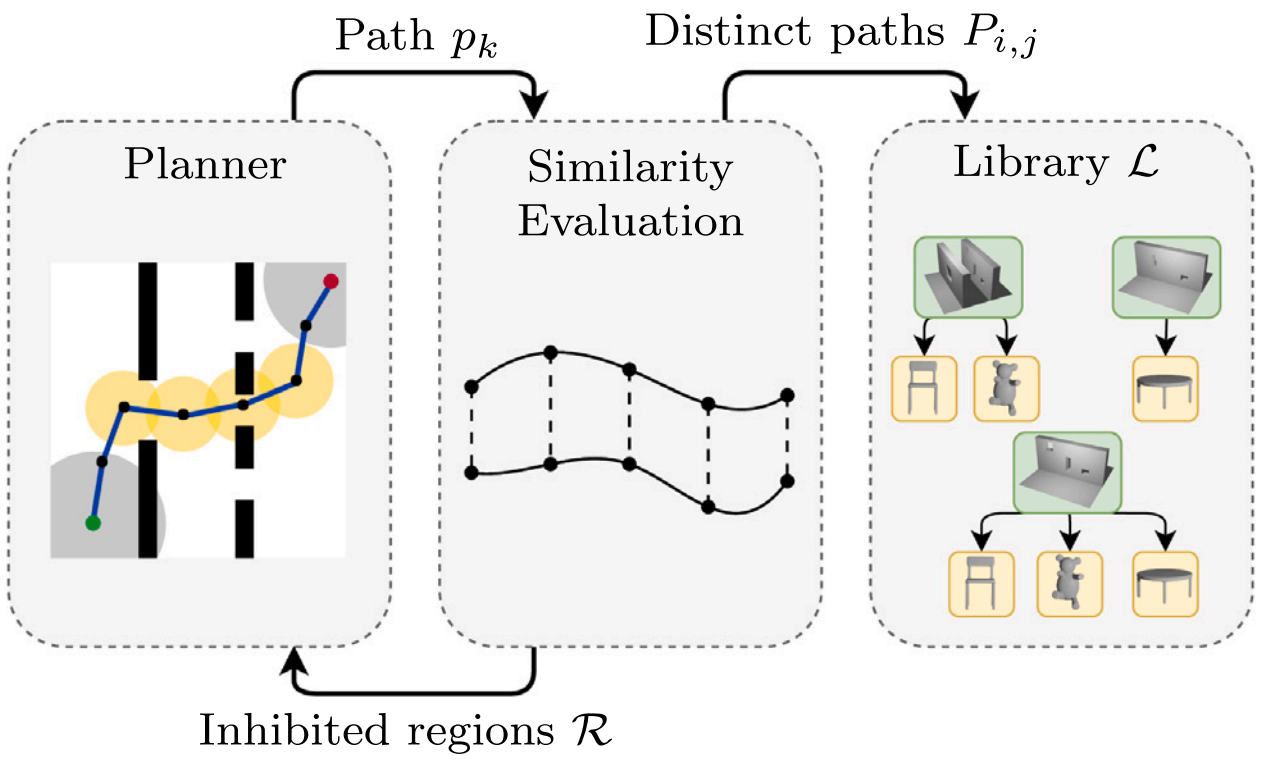

**Fig. 2.** The preparation phase of the proposed RRT-LIB algorithm. The planner iteratively finds multiple paths through the environment. Using the already-found paths as input to the planner increases the probability of finding distinct paths. After enough distinct paths are found, the planning is terminated, and the paths are saved in the library.

**Algorithm 1:** RRT-LIB preparation phase: finding paths for template object $\mathcal{O}_i$ in environment $\mathcal{W}_j$

**Input:** Template object $\mathcal{O}_i$ and its scaled-down variant $o_i$, workspace $\mathcal{W}_j$, start and goal configuration $q_{\text{start}}, q_{\text{goal}} \in \mathcal{C}_{\text{free}}$ in the configuration space formed by $o_i$ and $\mathcal{W}_j$, library of paths $\mathcal{L}$
**Params:** Minimal similarity of paths $d_{\min}$, Diversity patience $n_{\text{diversity_patience}}$, safe distance $d_{\text{safe}}$,
**Output:** Library of paths $\mathcal{L}$ updated by new paths $P_{i,j}$

$P_{i,j} = \emptyset$ // Generated paths
$\mathcal{R} = \emptyset$ // Inhibited regions
$n_{\text{same}} = 0$ // Successive non-distinct paths found
$k = 1$
**while** $n_{same} < n_{diversity_patience}$ **do**
  // Avoid the inhibited regions (Alg. 2 in [13])
  $p_k$ = RRT-IR($q_{\text{start}}$, $q_{\text{goal}}$, $\mathcal{G} = \emptyset$, $\mathcal{R}$)
  **if** $p_k \neq \{\}$ **then**
    **if** $\varrho_s(p_k, P_{i,j}) > d_{min}$ **then** // Eq. (2)
      // Add $p_k$ to $P_{i,j}$ only if distinct enough
      $P_{i,j} = P_{i,j} \cup \{p_k\}$
      $n_{\text{same}} = 0$
    **else**
      $n_{\text{same}} = n_{\text{same}} + 1$
    // Update the inhibited regions with a subset of $p_k$
    $\mathcal{R} = \mathcal{R} \cup \{q \in p_k \mid \varrho_q(q, q_{\text{start}}) > d_{\text{safe}} \wedge \varrho_q(q, q_{\text{goal}}) > d_{\text{safe}}\}$
    $k = k + 1$
$\mathcal{L} = \mathcal{L} \cup \{P_{i,j}\}$ // Update the library
**return** $\mathcal{L}$

The original RRT-IR can find multiple similar paths, which could decrease the efficiency of the library since we want to store only distinct paths. There is also no clear indication when we should stop searching for new paths, and input from the user (i.e., the number of iterations) is required. However, the number of iterations needed is hard to predict for complex environments. We extend the original RRT-IR with path diversity tracking to solve this issue. The distance between a pair of paths $\varrho_p(p_1, p_2)$ is measured as

$$\varrho_p(p_1, p_2) = \frac{1}{|p_1|} \sum_{q_1 \in p_1} \varrho_q(q_1, q_2^*), \qquad q_2^* = \arg\min_{q_2 \in p_2} \varrho_q(q_1, q_2), \tag{1}$$

where $\varrho_q(q_1, q_2)$ is a metric function on $SE(3)$, i.e., $\varrho_q : SE(3) \times SE(3) \rightarrow \mathbb{R}$. The distance of a path $p$ from a set containing multiple paths $P = \{p_1, \ldots, p_n\}$ is then computed as

$$\varrho_s(p, P) = \min_{k=1,\ldots,n} \max\{\varrho_p(p, p_k), \varrho_p(p_k, p)\}. \tag{2}$$

In order for a path to be considered distinct from the paths already found, its distance to the set containing the paths needs to be higher than a specified threshold $d_{\min}$.

Tracking the diversity of the guiding paths comes with two advantages. Adding a nondistinct path to the set of guiding paths would not give us any new information. Removing such paths results in fewer guiding paths while preserving diversity. This improves the speed of the second phase, where the guiding paths are used to guide the path search. However, the nondistinct path is still added to the inhibited regions to enforce searching for new, distinct paths. The second advantage comes from the fact that searching for new guiding paths can be automatically stopped when the path diversity reaches a plateau, in contrast to having to manually specify the number of guiding paths prior to the search phase. In reality, stopping as soon as the first nondistinct path is found could lead to missing more complicated paths. Therefore, the search is terminated once no distinct path has been found in the last $n_{\text{diversity_patience}}$ steps.

### 3.2. Planning phase

The aim of the planning phase is to find a path from $q_{\text{start}}$ to $q_{\text{goal}}$ for a given query object as fast as possible, which is achieved by guided sampling along a suitable path from the library (depicted in Fig. 3). First, the most similar template object $\mathcal{O}_i$ to the query object $\mathcal{O}_q$ is found in the library $\mathcal{L}$. We use a method presented in [19] which utilizes genetic algorithms to find the correspondence map between two meshes, minimizing the Average Isometric Distortion (summary presented in the SHREC'19 contest paper [60]). From the library, we select the object that minimizes this metric to the query object and retrieve the paths $p_k \in P_{i,j}$ computed for the template object along with correspondences $S_{q,i}$ between the query and the template object (pairs of mesh points, visualized in Fig. 4).

It can be expected that the query object $\mathcal{O}_q$ and the template object $\mathcal{O}_i$ do not share the same pose (e.g., a query desk object can be rotated, while the template desk object is not rotated, as in Fig. 5). In such a case, the approximate paths $P_{i,j}$ are not able to guide the search for the query objects. Therefore, the relative transformation of the query object to the template object is found using the Iterative Closest Point method (ICP) [20]. The results of ICP are the relative translation $t$ and rotation $R$. Then, the approximate solutions $P_{i,j}$ are transformed accordingly, i.e., all the waypoints are rotated and translated (Fig. 5). ICP is primarily used for correcting small transformations, and it is recommended to provide the algorithm with an initial guess of the correspondences. For that, we extend the original algorithm by using the correspondences $S_{q,i}$ retrieved during the similarity evaluation. These correspondences are computed using the method from [19], which identifies geometric similarities between meshes and can successfully establish matches even under large transformations between the query and the objects in the library. This initial estimate then enables the ICP algorithm to efficiently find a transformation between the objects. The extended ICP algorithm is outlined in Alg. 2. By adding the ICP transformation into the planner, we get the RRT-LIB planning phase outlined in Alg. 3.

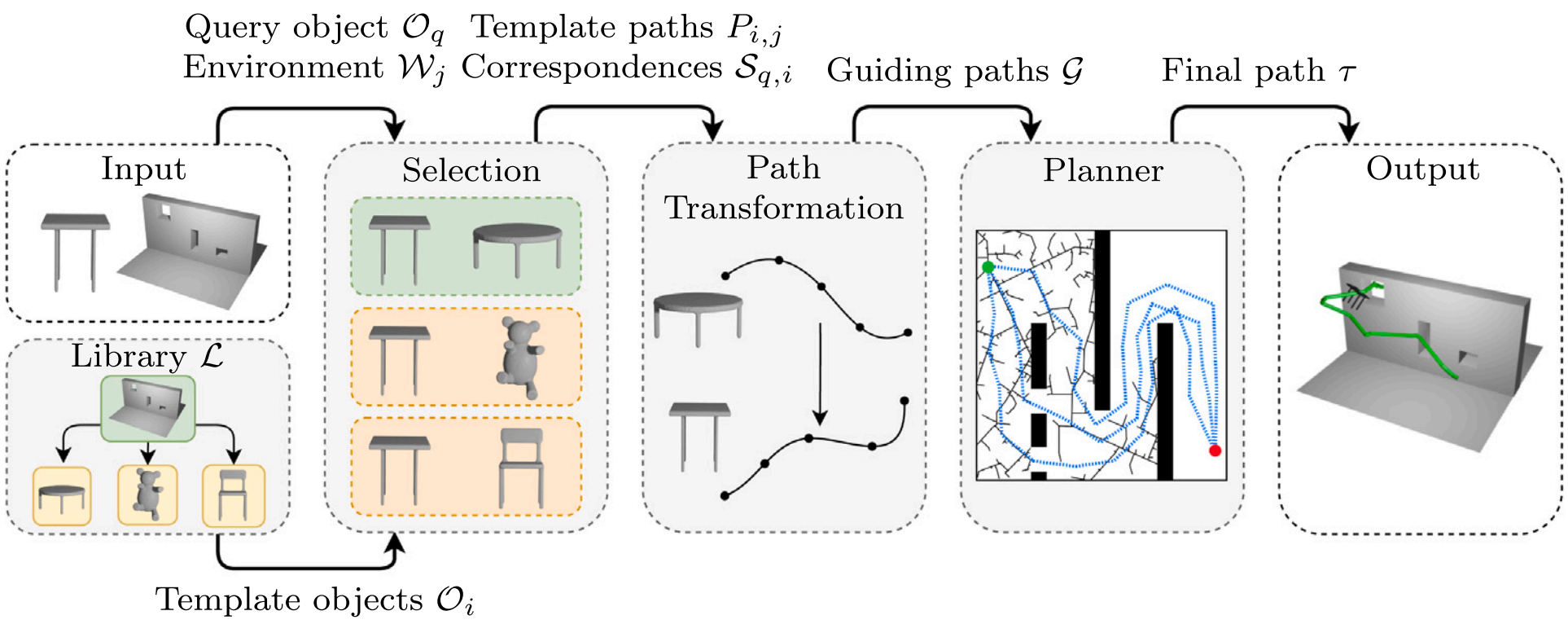

**Fig. 3.** The planning phase of the proposed RRT-LIB algorithm. The input consists of a manipulated object and an environment. Based on the inputs, paths computed for the most similar object in the library are loaded. After transforming the paths to account for possibly different positions of the manipulated object and the object from the library, we gain paths hinting at the possible paths through the environment. Increasing the sampling rate along these guiding paths should increase the probability of finding a solution.

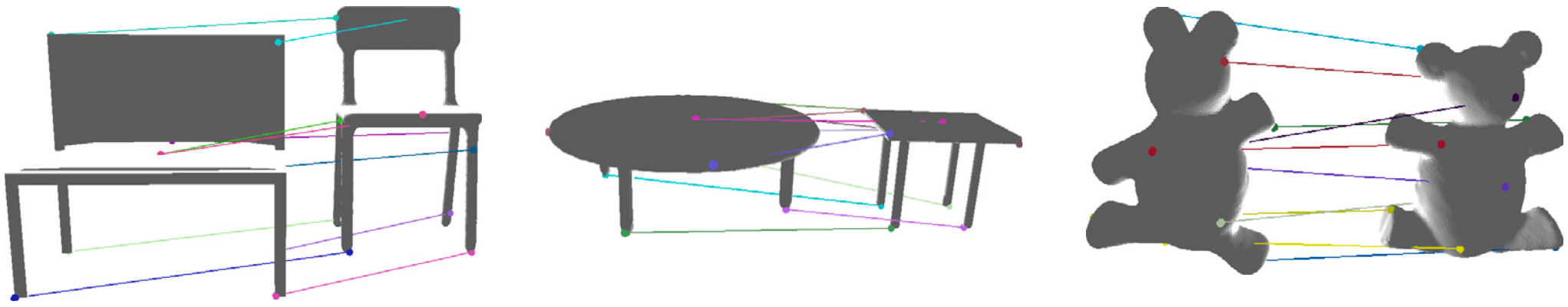

(a) chair (b) desk (c) teddy

**Fig. 4.** For every query object (left on each image), the most similar object from the library is selected (right on each image). The correspondences between the objects (illustrated by the colored lines) are used as the initial guess in the ICP algorithm. (For interpretation of the references to color in this figure legend, the reader is referred to the web version of this article.)

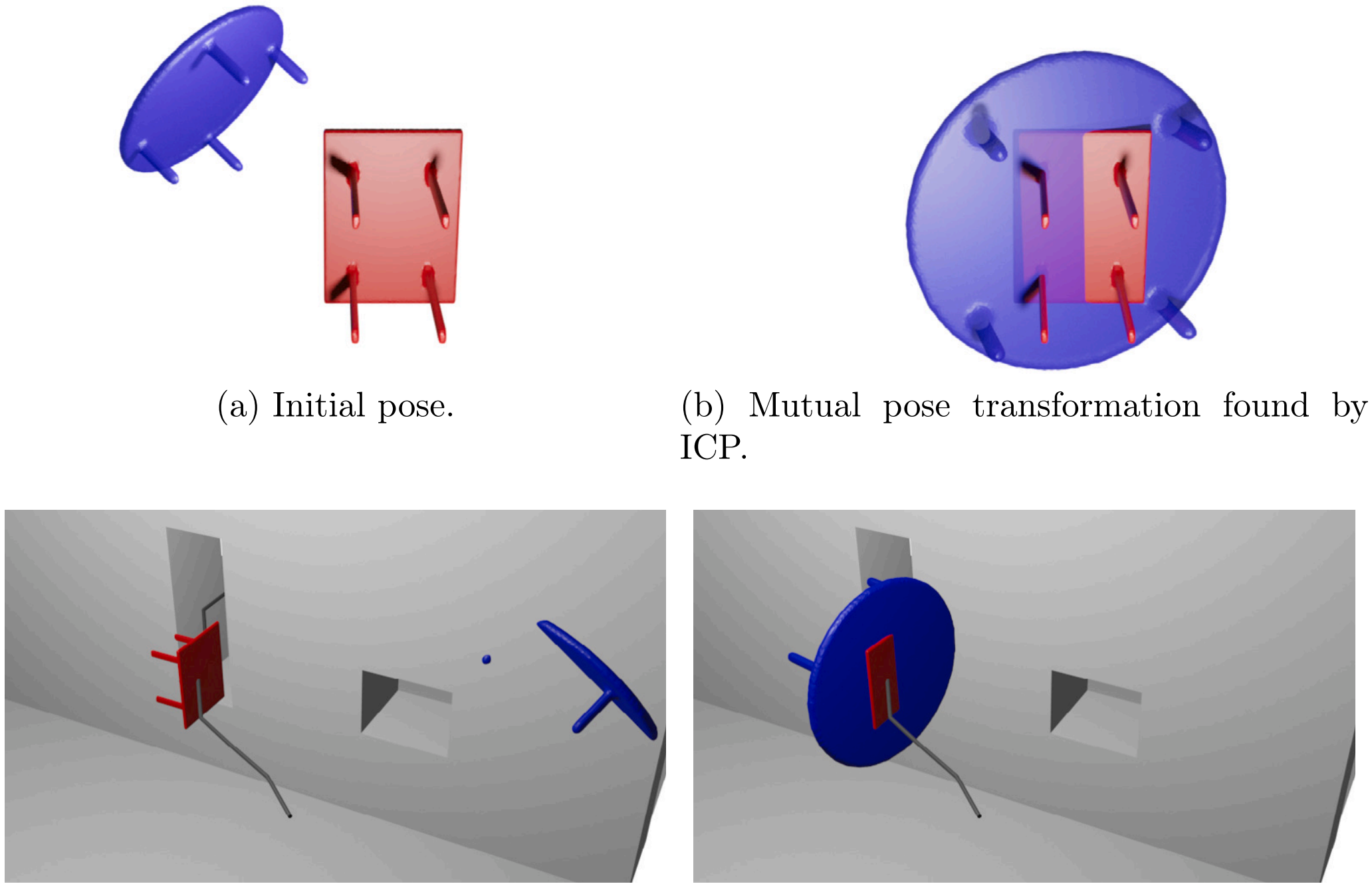

(a) Initial pose. (b) Mutual pose transformation found by ICP.

(c) Initial pose. (d) After the transformation is applied.

**Fig. 5.** When the mutual position and rotation of the guiding object (red) and the similar object (blue) are not taken into account, the guiding paths become useless because they guide the similar object through the wall instead of the middle window (Fig. 5(c)). After the guiding paths are transformed using the transformation obtained from ICP, planning along them becomes viable (Fig. 5(d)). (For interpretation of the references to color in this figure legend, the reader is referred to the web version of this article.)

**Algorithm 2:** ICP with an initial guess

**Input:** Source mesh $A = (a_1, a_2, \ldots, a_n)$,
Target mesh $B = (b_1, b_2, \ldots, b_m)$,
Initial correspondences $S = ((a_{s_1}, b_{s_1}), (a_{s_2}, b_{s_2}), \ldots, (a_{s_l}, b_{s_l}))$
**Params:** Maximum iterations $K$, Minimal error $\varepsilon_{\min}$
**Output:** Rotation matrix $R^*$, Translation vector $t^*$

$R^*, t^* = \min_{R,t} \sum_{i=1}^{l} \|b_{s_i} - (Ra_{s_i} + t)\|^2$ // Initial guess
$\varepsilon = \sum_{i=1}^{l} \|b_{s_i} - (R^* a_{s_i} + t^*)\|^2$
$k = 0$
**while** $k < K$ ***and*** $\varepsilon > \varepsilon_{min}$ **do**
  **for** $i = 1, \ldots, n$ **do** // Nearest vertex to
    $b'_i = nearest_point(a_i, B)$ // each vertex in source
  $R^*, t^* = \min_{R,t} \sum_{i=1}^{n} \|b'_i - (Ra_i + t)\|^2$
  $\varepsilon = \sum_{i=1}^{n} \|b'_i - (R^* a_i + t^*)\|^2$
  $k = k + 1$
**return** $R^*, t^*$

**Algorithm 3:** RRT-LIB planning phase

**Input:** Query object $\mathcal{O}_q$, Workspace $\mathcal{W}_j$, Configurations $q_{\text{start}}$, $q_{\text{goal}} \in \mathcal{C}_{\text{free}}$ in the configuration space formed by $\mathcal{O}_q$ and $\mathcal{W}_j$, Library of paths $\mathcal{L}$
**Output:** Path $\tau$ from $q_{\text{start}}$ to $q_{\text{goal}}$ or empty list

// Identify guiding object from $\mathcal{L}$
$\mathcal{O}_i, S_{q,i}, P_{i,j} = \mathcal{L}.query(\mathcal{O}_q, \mathcal{W}_j)$
// Compute mutual transformation (Alg. 2)
R, t = ICP($\mathcal{O}_q$, $\mathcal{O}_i$, $S_{q,i}$)
// Transform the library paths
$\mathcal{G} = transform_paths(P_{i,j}$, R, t)
// Plan along guiding paths $\mathcal{G}$ (Alg. 2 in [13])
$\tau$ = RRT-IR($q_{\text{start}}$, $q_{\text{goal}}$, $\mathcal{G}$, $\mathcal{R} = \emptyset$)
**return** $\tau$

## 4. Results

To compare the performance of our RRT-LIB planner with other planners, the Open Motion Planning Library (OMPL) [21] and its benchmarking suite [61] were used. The compared algorithms are listed in Table 1. The parameter settings for our RRT-LIB planner are listed in Table 2, with the default configuration used for all other planners. For each object and map pair, a benchmark test is performed, consisting of 50 runs with a 2-minute time limit each. Our primary criterion for comparison is the time required to find a solution — therefore, each run is stopped as soon as a feasible solution is found. The calculations are carried out on a computing grid *MetaCentrum*.[1] Each node runs a *Debian10* system with one thread on *CPU Intel Xeon Gold 5120* and 8 GB of RAM.

### 4.1. Preparation phase

All implemented methods are tested using objects from the PSB dataset [69] and manually created maps. Objects and maps are represented as triangulated meshes (each mesh has a few thousand triangles). To make the results comparable, each object is scaled so that its bounding box fits into a cube with an edge size of 2 map units. Scaling the objects does not affect our method of selecting the guiding object and correspondences, as shown in Appendix A. Three object categories (desks, chairs, and teddy bears) and two maps ($\text{map}_1$ and $\text{map}_2$) were chosen for the benchmarks. These maps were selected because they offer multiple distinct paths through the wall, and the window dimensions are designed to make the problem challenging enough but still solvable.

**Table 1**
Planners used in the benchmark.

| Abbreviation | Name |
|---|---|
| **RRT-LIB** | **RRT with a Library of Paths** |
| RRT | Rapidly-Exploring Random Trees [5] |
| RRTConnect | Bidirectional RRT [62] |
| Lazy RRT | Lazy vertex and edge evaluation RRT [63] |
| LazyPRM | Lazy vertex and edge evaluation PRM [64] |
| KPIECE | Kinematic Planning by Interior-Exterior Cell Exploration [65] |
| BKPIECE | Bidirectional KPIECE [65] |
| LBKPIECE | Lazy vertex and edge evaluation KPIECE [65] |
| EST | Expansive Space Trees [66] |
| BiEST | Bidirectional EST [66] |
| SBL | Single-query Bi-directional Lazy collision checking planner [67] |
| STRIDE | Search Tree with Resolution Independent Density Estimation [68] |

**Table 2**
RRT-LIB planner parameter values.

| Alg. 1 | | Alg. 2 | | RRT-IR | |
|---|---|---|---|---|---|
| Param. | Value | Param. | Value | Param. | Value |
| $d_{\min}$ | 1.20 | $N$ | 15 | $p_{\text{goal}}$ | 0.05 |
| $n_{\text{diversity_patience}}$ | 20 | $\varepsilon_{\min}$ | $10^{-10}$ | $p_{\text{bias}}$ | 0.80 |
| $d_{\text{safe}}$ | 0.80 | | | $d_{\text{guide}}$ | 0.50 |
| | | | | $d_{\text{inhibited}}$ | 1.20 |
| | | | | $d_{\text{safe}}$ | 0.80 |

From each object category (desk, chair, and teddy), one object is selected as the template. For scaled-down[2] versions of the objects, the template paths are computed in the two maps and saved to the library (visualized in Fig. 6). Our proposed method of generating template paths is able to find distinct paths quite well. However, no path was found through the smallest window in Fig. 6(b) and the middle window in Fig. 6(f). The template path generation takes approximately 20 s per object and map.

### 4.2. Computational comparison with the baseline algorithms

In each benchmark scenario, a query object and a map are given as inputs to the planners. At first, our method selects the most similar object from the library and computes the mutual correspondences. The selection of the most similar object takes, on average, one second per object in the library — therefore, in our case, it adds approximately 3 s to each planning task. In all cases, the shape correspondence algorithm successfully selected the template object belonging to the same class as the query object (Fig. 4). The paths computed for the template object are retrieved from the library, and a transformation between the template object and the query object is found by ICP to ensure the objects have a similar position and rotation. Finding the transformation using ICP is not time demanding, and it takes 140 ms on average to compute for our data and computers used. After the transformation is applied to the paths from the library, they are used as the guiding paths for the planner. If no solution is found after two minutes (this time includes the similarity evaluation and ICP for our RRT-LIB planner), the run is terminated.

A representative subset of the benchmark results is presented in Figs. 7 and 9, with a bar graph representing the success rate of each planner (in how many of the ten runs a path was successfully found) and a box graph containing the time needed to find a solution (capped at the 2-minute threshold). In the easier (i.e., the objects can pass through multiple windows and do not require complex maneuvers to do so) benchmarks (Fig. 7), our planner achieves a 100% success rate, along with some other planners. However, considering the time added by generating the template paths in the *preparation phase* (not included in the timing), selecting the template object from the library, and finding the mutual transformation using ICP in the *planning phase* (∼3 s, included in the timing), using one of the other planners would still be faster in some cases. A detailed view of the results is shown in Fig. 8. In the first test, our planner solves the task in 3.5 s on average, whereas the RRT and RRTConnect need 2.2 s and 1.4 s on average, respectively. Here, the main contribution to the time needed by the RRT-LIB planner is the similarity evaluation (∼3 s). In the second and third tests, the advantage of retrieving the guiding paths for a similar object and transforming them for the given task starts to show. In the second test, the average runtime of our RRT-LIB planner is 3.4 s — a 50% improvement over the second best algorithm in the test (RRTConnect with an average of 6.8 s). In the third test, the average runtime of our RRT-LIB planner is 3.6 s — an approximately 85% improvement over the second best algorithm in the test (RRT with an average of 24.3 s).

[1] https://www.metacentrum.cz/cs/.

[2] Objects were scaled-down to 40% of their original size.

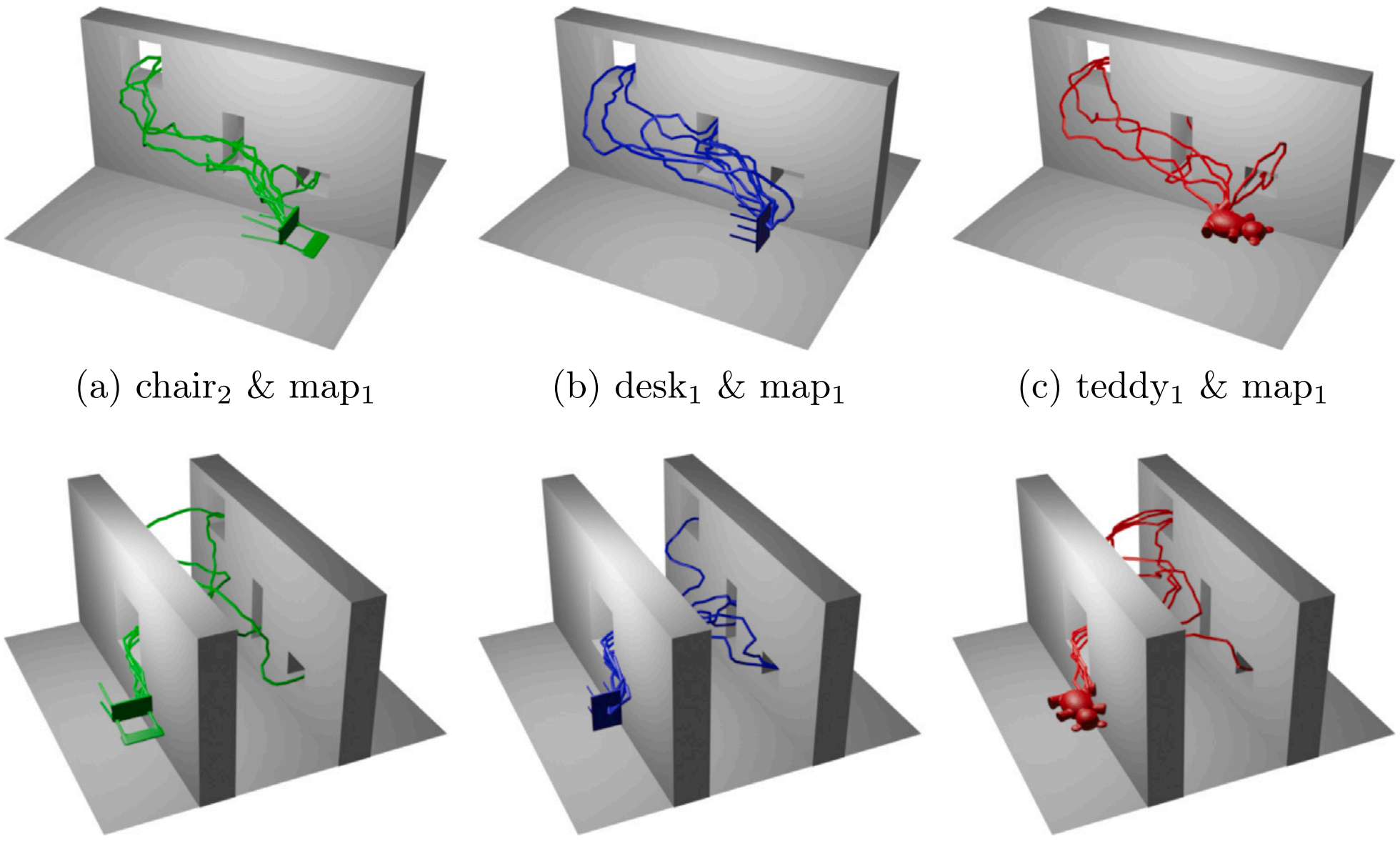

(a) $chair_2$ & $map_1$ (b) $desk_1$ & $map_1$ (c) $teddy_1$ & $map_1$

(d) $chair_2$ & $map_2$ (e) $desk_1$ & $map_2$ (f) $teddy_1$ & $map_2$

**Fig. 6.** Guiding paths for the three guiding objects and two maps. Generated as a part of the RRT-LIB preparation phase.

Tests depicted in Fig. 9 are substantially more challenging (i.e., the objects can only pass through one of the windows, and complex maneuvering is needed) and show the real advantage of using the guiding paths retrieved from the library and transformed to respect the manipulated object. Our planner is still able to find a solution in most of the runs, while the other planners fail every time. Specifically, in the first and third scenario, our RRT-LIB planner retains the 100% success rate, finding a solution in 3.8 s and 10.2 s on average, respectively. In the third scenario, our planner failed to find a solution in two of the 50 runs, while the other planners were unable to provide any solution. The fact that none of the other tested planners produced a solution within the 2-minute time limit (in 50 trials) demonstrates the superior performance of the proposed RRT-LIB, which achieves a 96% success rate. A higher success rate could potentially be obtained by either increasing the allowed time limit or tuning the parameters, but this is beyond the scope of this paper.

To demonstrate the transferability of the information contained in the library, we concluded another experiment with a much larger object. The object's minimal 3D bounding box is considerably larger than any of the windows, and a precise manipulation is needed to pass the narrow passage, as shown in Fig. 10. Our RRT-LIB planner is still able to solve such problem with a 100% success rate in the imposed 2-minute limit, while the other planners fail every time.

## 5. Discussion

The overhead introduced by the various steps in the pipeline (i.e., computing similarity between objects, finding their transformation using ICP, and planning) must be considered when evaluating the usefulness of the RRT-LIB planner for a specific task. The most time-consuming part of each query is the computation of similarity between the query and the objects in the library, which is performed using the method in [19] and takes approximately 1 s per object (the objects used in the experiments are described by thousands of triangles). The evaluation of similarity can be accelerated by employing alternative shape-matching methods [70], possibly combined with mesh simplification [71]. Mesh simplification would also speed up the ICP step; however, in our experiments, ICP takes only 140 ms per query, which is negligible compared to the time required for computing the similarity.

When the planning problem is easy to solve, the time required to select the template object and compute the mutual transformation can be considerably longer than the planning itself. Although the planner finds the solution almost immediately using the guiding paths, no significant time is saved compared to a standard RRT planner. Therefore, applying RRT-LIB in simple scenarios (e.g., convex robots navigating among widely spaced obstacles without narrow passages) provides little to no advantage.

In contrast, the benefit of using guiding paths computed for similar objects becomes apparent in challenging scenarios, where narrow passages and complex obstacles make planning significantly more difficult. In these cases, RRT-LIB substantially accelerates the search. Moreover, it successfully finds paths even when the other planners fail. However, the library needs to contain computed paths, and the computation of guiding paths in the RRT-LIB preparation phase requires a considerable amount of time. This requirement is similar to the learning-based path-planning approaches, which require significant training time [54–56]. Making such an effort is reasonable only when planning for similar objects in the same environment repeatedly. Suppose that we know that the planning will be performed multiple times with a defined set of static environments and object classes. In that case, we can reuse the experience gained by prior planning, and devoting the time to compute paths to store in the library will prove advantageous in the long run. However, when the planning task is unique and will not be repeated,

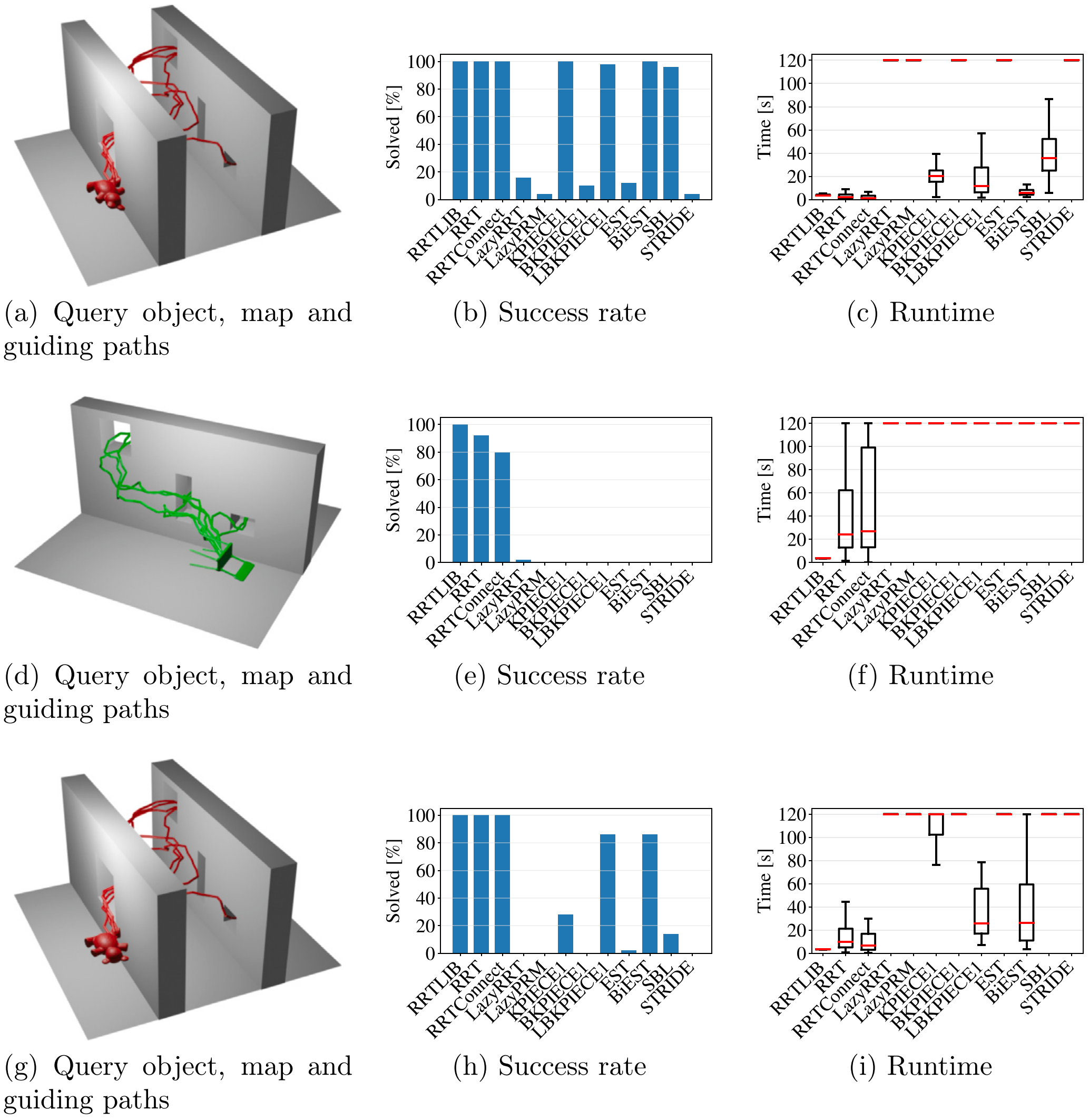

(a) Query object, map and guiding paths (b) Success rate (c) Runtime

(d) Query object, map and guiding paths (e) Success rate (f) Runtime

(g) Query object, map and guiding paths (h) Success rate (i) Runtime

**Fig. 7.** Easier benchmarks results.

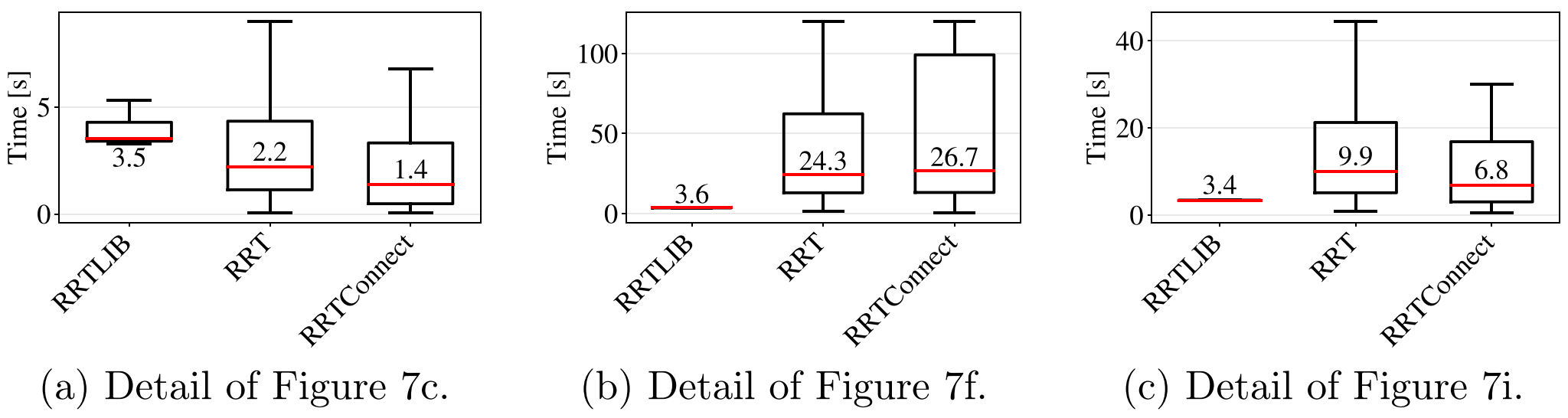

(a) Detail of Figure 7c. (b) Detail of Figure 7f. (c) Detail of Figure 7i.

**Fig. 8.** Detailed view of three benchmark results with median values shown. The runtime of our RRT-LIB planner is compared to the standard RRT [5] and its bidirectional variant (RRTConnect) [62].

one should also consider other approaches when choosing the right planner.

Alternatively, RRT-LIB could be used even with an empty database $\mathcal{L}$ and update the database after new paths are found. In such a case, the planner (Alg. 3) would start with no guiding paths ($\mathcal{G} = \emptyset$) and would perform a standard RRT search. After finding a path, the planner would store the object as a template and the path as a guiding path in the database. During subsequent planning tasks, each newly found path would be stored only if it was distinct enough from the already stored paths (as in Alg. 1, line 8). Updating the library by the latest plan is common also in related works [44,49–51]. However, these works assume only one type of robot, so the new paths are simply added to the library. In RRT-LIB, multiple objects (robots) are assumed — if the object was deemed too different (which could be implemented trivially by thresholding the similarity metric used for similarity evaluation), a new template object would be created instead.

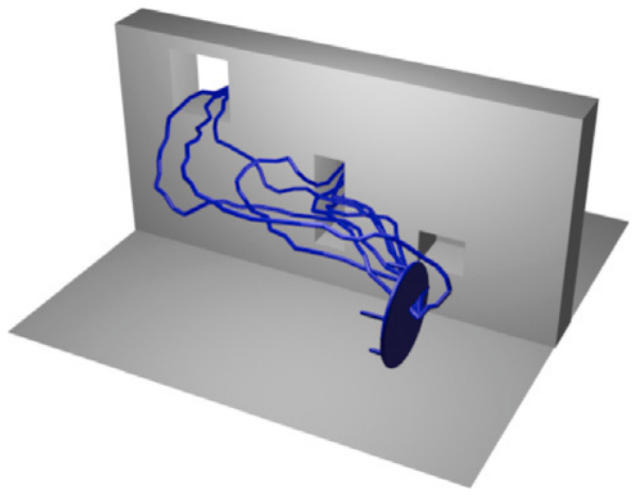

(a) Query object, map and guiding paths

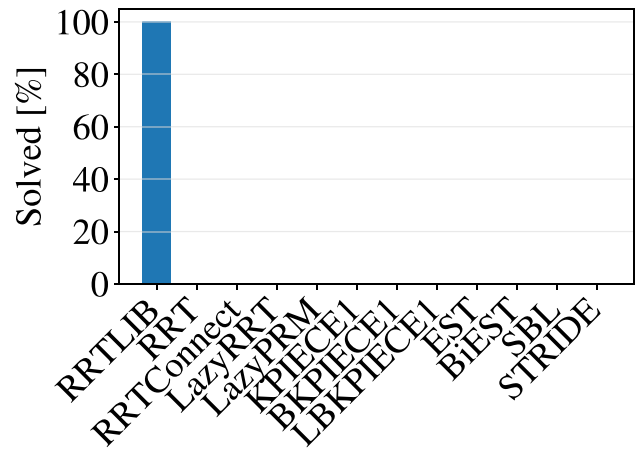

(b) Success rate

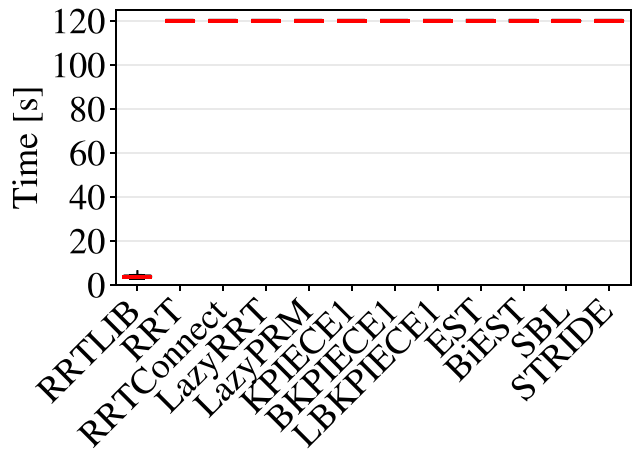

(c) Runtime

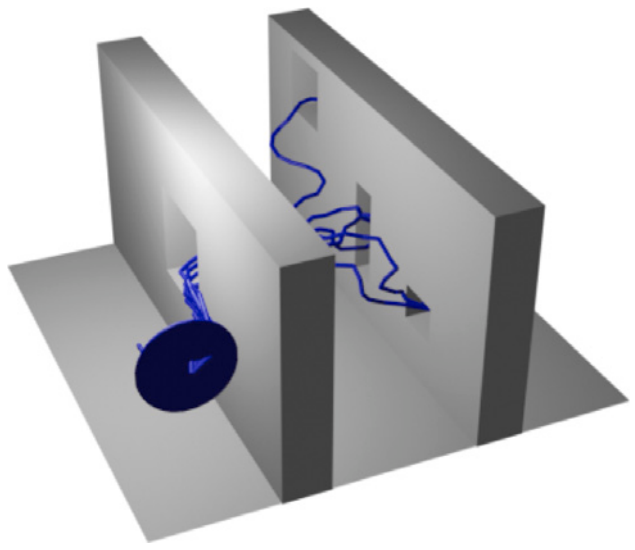

(d) Query object, map and guiding paths

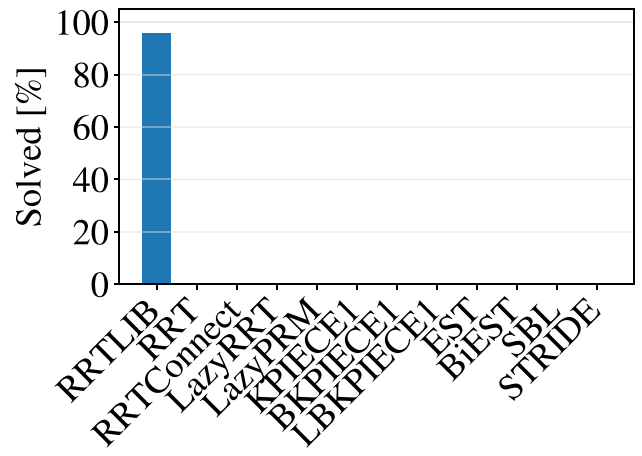

(e) Success rate

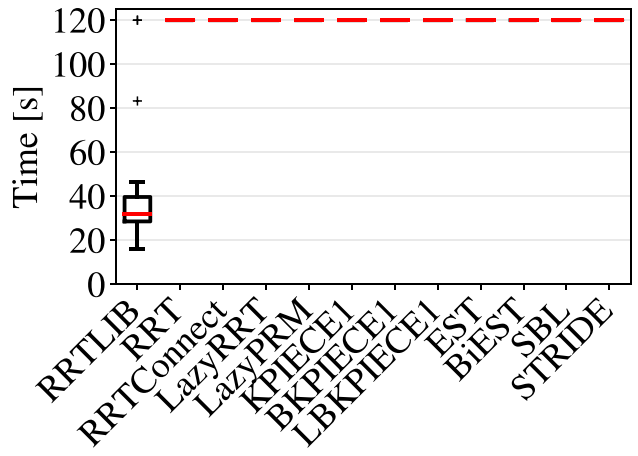

(f) Runtime

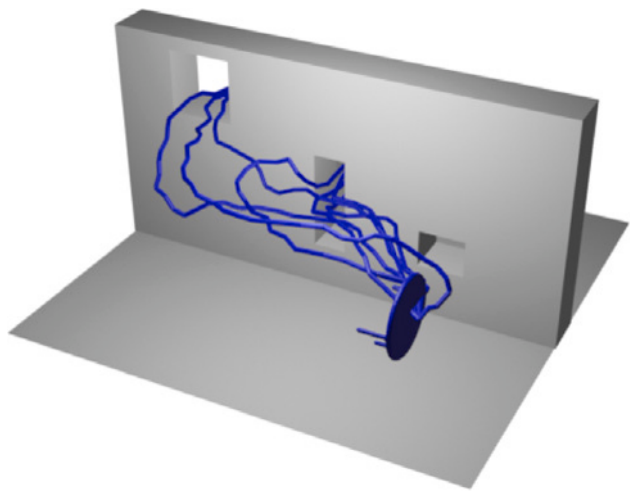

(g) Query object, map and guiding paths

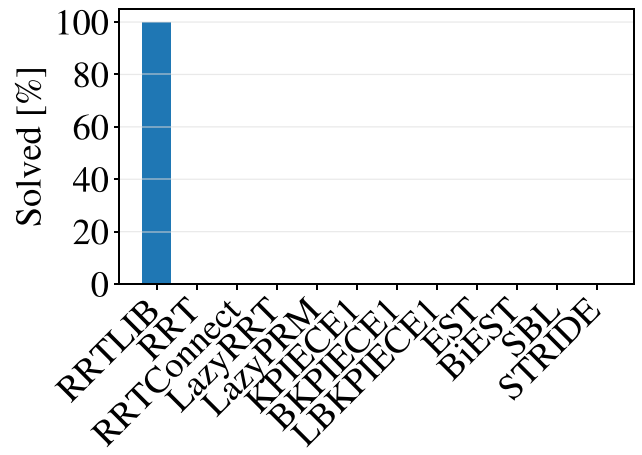

(h) Success rate

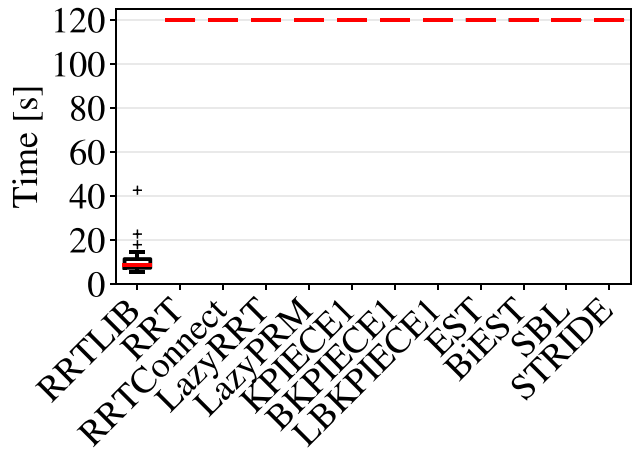

(i) Runtime

**Fig. 9.** Challenging benchmarks results.

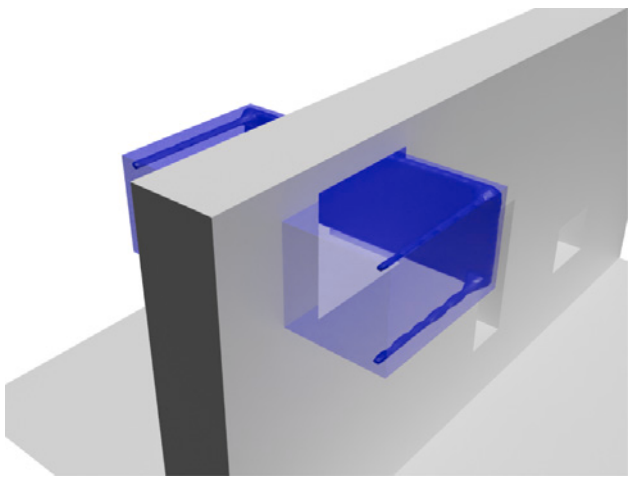

(a) Query object and its minimal bounding box

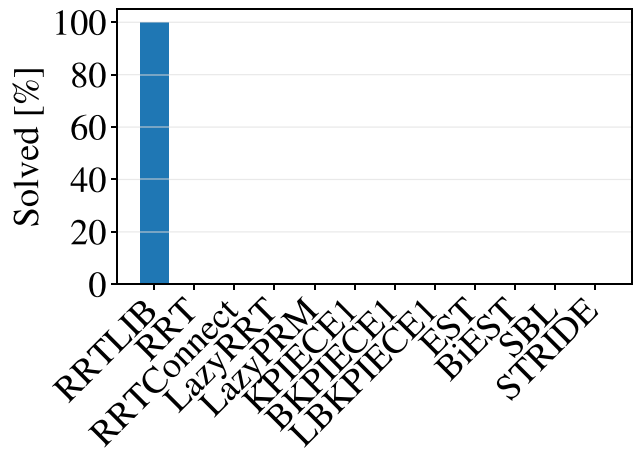

(b) Success rate

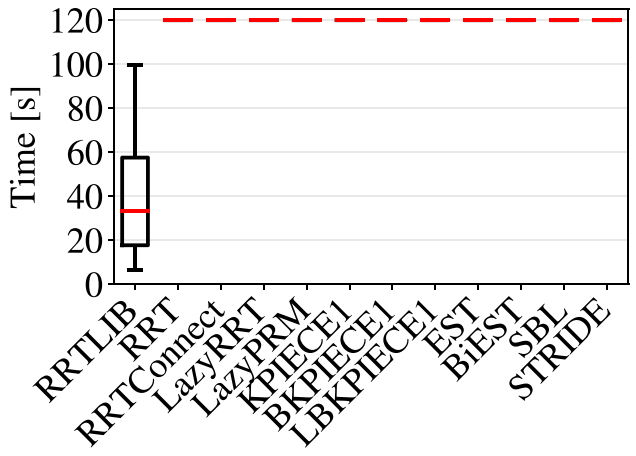

(c) Runtime

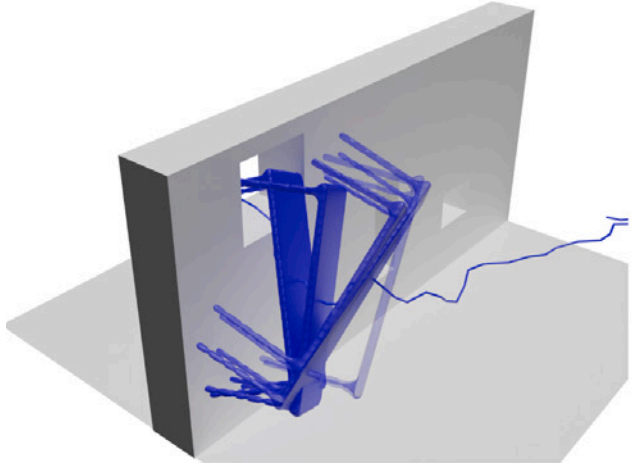

(d) RRT-LIB solution: Entry

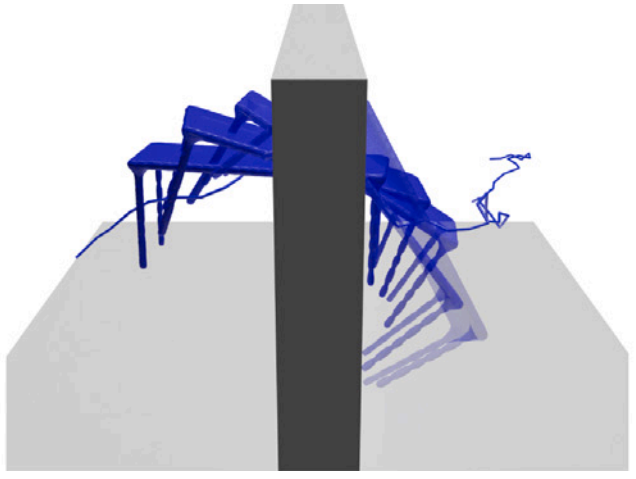

(e) RRT-LIB solution: Narrow Passage

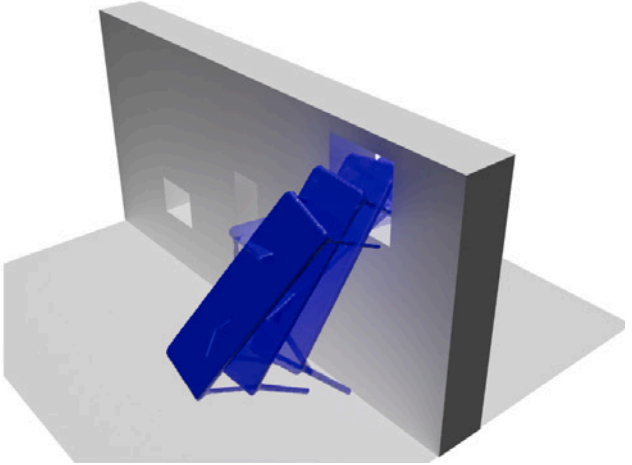

(f) RRT-LIB solution: Exit

**Fig. 10.** The proposed RRT-LIB planner is able to efficiently solve a problem where passing through the narrow passage requires precise manipulation.

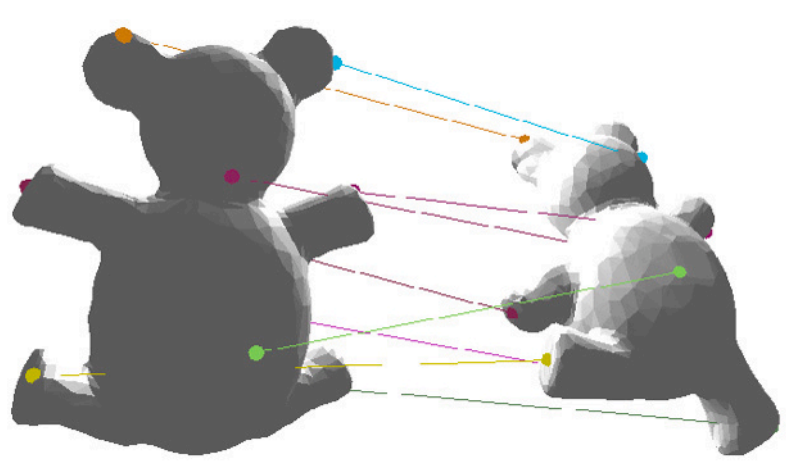

(a) Target rotated by −45° around x-axis.

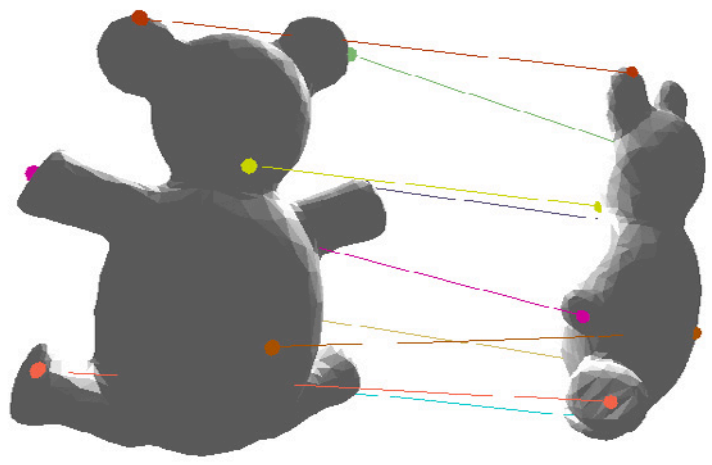

(b) Target rotated by 45° around y-axis.

**Fig. A.11.** The rotation of the object does not change the found correspondences.

## 6. Conclusion and outlook

This work presented RRT-LIB, an algorithm that aims to improve the efficiency of sampling-based planners by creating a library with already computed paths in an environment with narrow passages. We generate distinct paths for multiple object classes using Rapidly-Exploring Random Trees with Inhibited Regions (RRT-IR). These paths are filtered based on their similarity using our proposed method, and a diverse subset is saved in the library. During a query, the object most similar to the given manipulated object is found in the library using shape-matching methods based on Genetic Algorithms. To ensure that the library object and the manipulated object are positioned similarly in the coordinate frame, a transformation between them is found by the ICP algorithm. Finally, the transformed paths are used to guide the planner through the narrow passages contained in the environment.

We compared our planner to other state-of-the-art path planning methods using an open-source planning library OMPL [21]. The benchmarks demonstrated that having multiple precomputed paths through the environment for a template object can substantially decrease the time needed when planning for a new similarly-shaped object. Our RRT-LIB planner was able to outperform the other state-of-the-art planners, with a 50% and 85% lower runtime in two of our tests compared to the second-best planner in each test. More importantly, our planner is able to find paths even in cases where the other planners fail. This was shown in three test scenarios, where the other planners were unable to find any solution in 50 test runs, while our algorithm found the solution in 100% runs in two of the tests and in 96% runs in another test scenario.

RRT-LIB offers a modular pipeline (with its implementation available as open-source), whose reliability is shown in this work, even in scenarios where a precise manipulation is required. Most notably, this paper proposes a method to transform precomputed paths using correspondences from similarity evaluation, and empirically demonstrates its viability and robustness. With this foundation in mind, multiple prospective research directions are thereby open for future work, as many individual problems remained beyond the scope of this work.

We have limited ourselves to planning problems with the same start and goal configuration for each object, allowing us to use the whole precomputed path. Even though this assumption is not restrictive thanks to the proposed path transformation, decomposition of the library paths into smaller parts could be a viable extension. The path transformation presented in this work could be easily applied to the smaller parts, thereby generalizing the capabilities of the RRT-LIB planner. Most notably, an emphasis on the parts traversing the narrow passages could be made, reducing the focus on large open portions of the free space, where planning is easy. However, even the identification of narrow passages is a non-trivial problem, as a narrow passage is specific to both the environment and the manipulated object.

Another direction would be the implementation and evaluation of a more complex path-filtering method, since even two homotopic paths can be declared distinct if they are far enough in the free space. Among the more advanced methods for filtering paths based on their path through the environment is topological clustering [72,73] or Maximal Path Diversity Pruning [74]. This could lead to improved results in more types of environments.

A combination of such path decomposition and more complex filtering might allow efficient large-scale library usage and improve the efficiency of the proposed method. Moreover, decomposing the path into smaller portions might prove helpful in real-world trials, particularly when the manipulated object is displaced during the path execution (e.g., by a human operator or due to dynamic object avoidance).

## CRediT authorship contribution statement

**Michal Minařík:** Writing – review & editing, Writing – original draft, Visualization, Validation, Software, Methodology, Investigation, Formal analysis, Data curation, Conceptualization. **Vojtěch Vonásek:** Writing – review & editing, Supervision, Resources, Project administration, Investigation, Funding acquisition, Formal analysis, Conceptualization. **Robert Pěnička:** Writing – review & editing, Supervision, Project administration, Investigation, Funding acquisition.

## Acknowledgments

This work has been supported by the Czech Science Foundation (GAČR) under project No. 24-12360S, and by the European Union under the project Robotics and advanced industrial production (reg. no. CZ.02.01.01/00/22_008/0004590). Computational resources were provided by the e-INFRA CZ project (ID:90254), supported by the Ministry of Education, Youth and Sports of the Czech Republic.

## Declaration of competing interest

The authors declare that they have no known competing financial interests or personal relationships that could have appeared to influence the work reported in this paper.

## Appendix A. Transformation invariance of shape matching

To examine how the selected method for shape matching reacts to different object transformations, we rotate and scale the target object while keeping the initial object unchanged. The algorithm is run for each transformation, and the correspondences are evaluated. The resulting correspondences found between the initial and the target object are not affected by either rotation (Fig. A.11) or scaling (Fig. A.12).

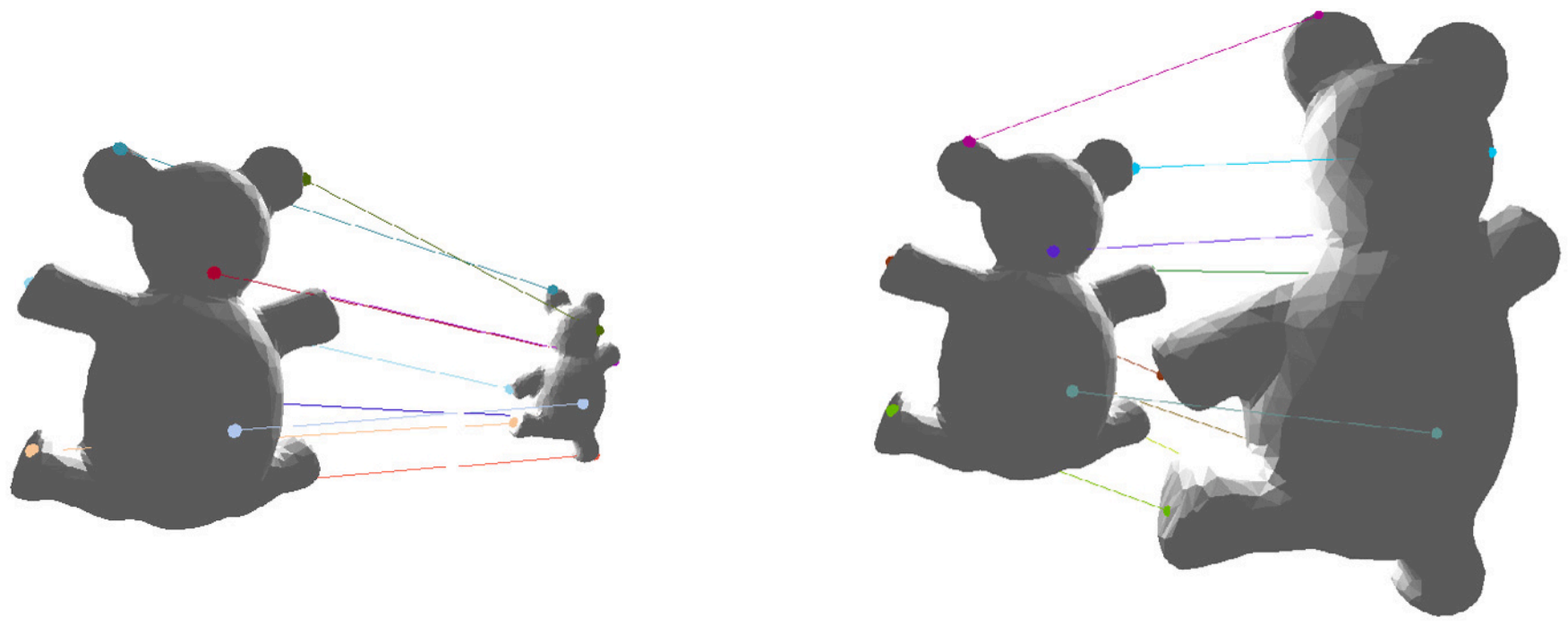

(a) Target scaled 0.5x (b) Target scaled 2x

**Fig. A.12.** The scale of the object does not change the found correspondences.

## Appendix B. Supplementary materials

**Code:** https://github.com/m-minarik/rrtlib
**Video:** https://youtu.be/1BTlWC742Aw

## Data availability

Data will be made available on request.

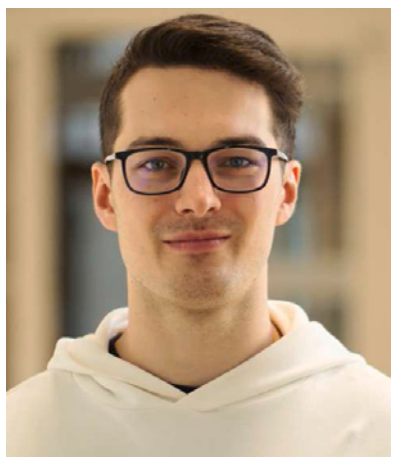

**Michal Minařík** is a postgraduate student in the Multi-Robot Systems (MRS) group at the Czech Technical University (CTU) in Prague. He graduated with honors in 2024 and was awarded the Dean's Award for his Master's Thesis. His interests are planning and control with a focus on agile Unmanned Aerial Vehicles, along with mathematical optimization and teaching, for which he received a Dean's Teaching Award for excellent teaching performance in 2023.

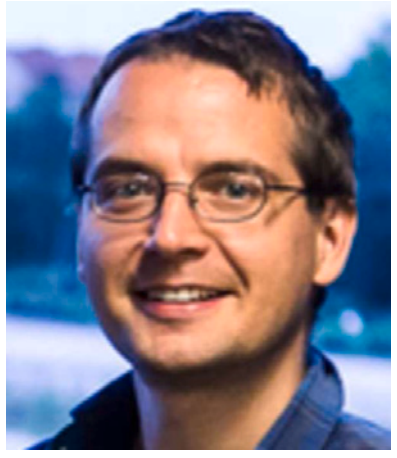

**Vojtěch Vonásek** obtained his Ph.D. degree from FEE, CTU in 2016. He spent his post-doc stay at KIT, Karlsruhe, Germany and TU-Berlin, Berlin, Germany. Currently, he is a post-doc at Multi-robot systems Group at FEE, CTU. His research interests are path and motion planning, physical simulations and the applications of these in bioinformatics.

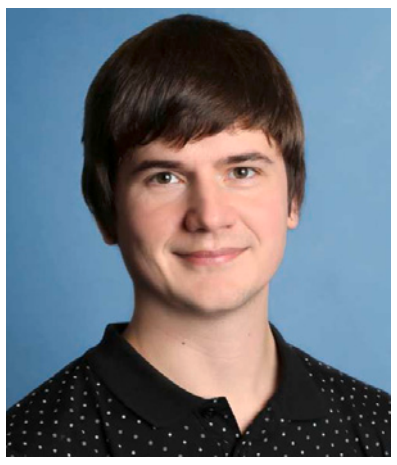

**Robert Pěnička** is currently a postdoc in the Multi-Robot Systems (MRS) group at the Czech Technical University (CTU) in Prague. He did his Ph.D. at the CTU in Prague in 2020 and was a postdoctoral researcher at the University of Zurich between 2020 and 2022 under the supervision of Professor Scaramuzza. From 2022 he is a research fellow at the CTU and part of the MRS. He focuses on high-level mission planning, trajectory planning, and control for Unmanned Aerial Vehicles (UAV). He won several awards for his doctoral thesis on multi-goal mission planning for UAVs including Dean's Prize, 2nd place in 2020 Werner von Siemens Award for best thesis in Industry 4.0 in the Czech Republic, 2nd place in Joseph Fourier Prize 2021 for the best doctoral thesis in computer sciences in the Czech Republic, and Antonin Svoboda Award 2020 for the best doctoral thesis from the Czech Society for Cybernetics and Informatics.